# HRCenterNet: An Anchorless Approach to Chinese Character Segmentation in Historical Documents

Chia-Wei Tang
Department of Computer Science
National Chengchi University
Taipei, Taiwan
106703054@g.nccu.edu.tw

Chao-Lin Liu
Department of Computer Science
National Chengchi University
Taipei, Taiwan
chaolin@g.nccu.edu.tw

Po-Sen Chiu
Department of Computer Science
National Chengchi University
Taipei, Taiwan
107753029@g.nccu.edu.tw

*Abstract*— The information provided by historical documents has always been indispensable in the transmission of human civilization, but it has also made these books susceptible to damage due to various factors. Thanks to recent technology, the automatic digitization of these documents are one of the quickest and most effective means of preservation. The main steps of automatic text digitization can be divided into two stages, mainly: character segmentation and character recognition, where the recognition results depend largely on the accuracy of segmentation. Therefore, in this study, we will only focus on the character segmentation of historical Chinese documents. In this research, we propose a model named HRCenterNet, which is combined with an anchorless object detection method and parallelized architecture. The MTHv2 dataset consists of over 3000 Chinese historical document images and over 1 million individual Chinese characters; with these enormous data, the segmentation capability of our model achieves IoU 0.81 on average with the best speed-accuracy trade-off compared to the others. Our source code is available at https://github.com/Tverous/HRCenterNet.

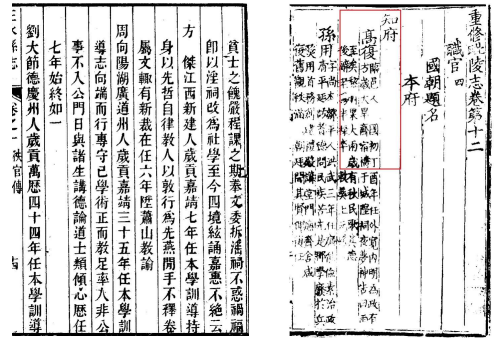

Fig 1. Different formatting between different documents. The document on the left is more monotonous, but the document on the right, in addition to features commonly found in Chinese historical documents, has different issues like various font sizes.

## I. INTRODUCTION

Since the establishment of writings, the transmission of civilizations has relied heavily on the contents of historical documents. Therefore, the preservation and interpretation of ancient books have always played a pivotal role in developing human civilizations. The modern writing system in the Chinese language has been in common use for hundreds of years, and the accumulation of ancient books and knowledge is not to be underestimated, so the conservation of these books has always been an important issue. The preservation of historical documents is not easy, and it is necessary to consider various factors that may affect the quality of the books. Factors such as climate and location have to be taken into account to ensure the integrity of books. Thus, the digitization of historical books has become a priority with the development of technology in recent years, but the digitization of historical documents requires many experts to achieve. In the Chinese writing system, different characters such as Chinese cursive or Chinese semi-cursive may not be fully understood by most modern people, and complicated and tedious human labelings are unavoidable in the digitization process. All of these reasons make the digitization of Chinese historical documents extremely difficult. If the documents can be digitized automatically with the help of modern technology, this will significantly reduce the cost and risk of the work. The current research process of digitizing Chinese historical documents can be divided into two stages, mainly: the Chinese character segmentation stage and the Chinese character recognition stage. In order to obtain accurate recognition results, it is necessary to obtain accurate results in Chinese character segmentation. As a result, in this study, with the success of deep learning on various topics recently, we propose a general and comprehensive solution with a parallel model architecture to the segmentation of Chinese characters from historical Chinese documents.

*A. A Brief Primer on the Contents of Chinese Historical Documents.*

To make this study more understandable, in this section, we will introduce the characteristic of Chinese characters, the written format of the Chinese historical documents, and the difficulties encountered in digitization. Unlike Western writing systems, where only a limited number of alphabets are used to form new words and phrases, each word and phrase in the Chinese language is made up of different symbols, each representing an individual meaning, and the small differences of the symbols can lead to completely different results. In other words, compared to the letters of the Western alphabet, the Chinese alphabet is overwhelmingly large and complex, which greatly complicates the digitalization of these documents. The arrangement of most Chinese historical documents follows the format based on top-down, right-to-left, vertical lines, as illustrated in Figure 1. This composition is advantageous for traditional image processing techniques; however, it is only feasible for specific forms. Other issues include:

*1) Various composition in Chinese Historical Documents*

In some cases, the size of the font or the formatting of the text is changed in the document to indicate special meaning or to supplement explanations, resulting in multiple words in the same row or column, as shown on the right-hand side of Figure 1. This makes it impossible for certain algorithms based on specific rules to work properly. Although there has been significant improvement in OCR lately with the success of deep learning, due to the complexity of Chinese historical documents and the lack of labeled data, there have been no breakthroughs.

## 2) Distortion in Historical Documents

For damaged documents, it is difficult to retrieve the text's information, as shown in Figure 2. This will make it difficult for the detector to detect the Chinese characters in the document.

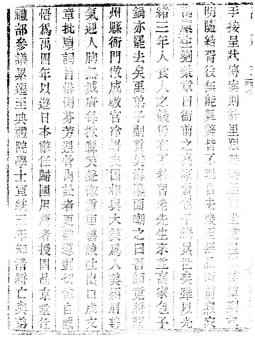

Fig 2. Damaged document or distortion in upstream image data can cause the content to be undetectable and unrecognizable.

### B. Object Detection in Character Segmentation

Due to the excellent performance of CNN(Convolutional Neural Network) in image processing recently, many object detection models based on this technique have been proposed [1-8, 13], most of which require the generation of multiple anchors[1] to generate ROIs(Region of Interest), which are then classified by other models to produce the final result. However, there are two major drawbacks to apply anchor-based models for character segmentation. The first one is the complexity of the computation. In most Chinese historical documents, the image of one page can contain more than 200 individual Chinese characters, which makes it very computationally intensive to use anchors to generate individual ROIs and then classify them separately. The second problem is that using an anchor-based model usually involves many hyperparameters, such as the number of anchors that need to be generated, the size of each anchor, the threshold value of ROI, and so on. Especially for the topic of exceptionally densely character detection, the related hyperparameters are very difficult to set, and as mentioned above, different document formats can not be fit in the same setting in most cases. This makes anchor-based models unsuitable for character segmentation in Chinese historical documents and is one of the reasons why past research has failed to make a breakthrough.

To overcome these anchor-based model problems, [6] proposes an object detection model that does not require anchors but represents each desired object with a set of keypoints[2] and predicts these keypoints directly from the model, which allows the object detection problem to be transformed into a standard keypoint detection problem without anchor intervention or other processing.

In addition to the above issues, another common difficulty with object detection is that it is hard to maintain high accuracy in preceding models of position-sensitive tasks such as character segmentation. To avoid distortion of the output from the original input, high-resolution representations are required. However, most of the models in the past followed the process of reducing the spatial size of the feature map step by step and then performing subsequent recognition or segmentation tasks by restoring directly from the low-resolution representations[10-13]. This approach is especially detrimental to tasks in a greater need for tiny features, especially to the subject of character segmentation from documents with very dense desired objects, because some important features may be lost during extraction. While the model proposed by [9] is able to extract high-level features and restore images from it while maintaining high-level representations so that relevant information about the task is not easily lost and subsequent results are more accurate.

To solve these predicaments, we propose a generalized Chinese character detection model, namely HRCenterNet with high accuracy inspired by [6-9]. For feature extraction, the parallelized model is implemented so that the global feature and local feature can be considered together, and high-level and low-level representations can interact with each other under multi-scale fusion. In the final segmentation result, unlike anchor-based models in the past, our model can achieve scale-aware and accurate segmentation.

We use the Chinese character labeled data MTHv2 provided by [14] as training and testing data to compare this model with other previous models for Chinese character segmentation. We achieve an IoU(Intersection over Union)[3] of 0.81 with the best speed-accuracy trade-off compared to the others.

## II. RELATED WORKS

In the past, the topic of using OCR to segment text from backgrounds has been widely explored, and two main techniques have been used: traditional image processing techniques and, more recently, deep neural network techniques using CNN in image processing. We will discuss them separately below.

### A. Traditional image processing techniques

Since most Chinese historical documents follow a fixed writing pattern, all characters are independent and separated from each other so that the characters can be divided into rows or columns by horizontal and vertical projections; the method based on morphology and histogram projection algorithm then becomes one of the suitable ways to segment Chinese characters from the documents. However, this method still encounters many bottlenecks, mainly due to the lack of generality and a fixed writing format restriction. A slight difference in the format in these documents will cause a significant difference in the definition of projection width and projection threshold, which increases the segmentation's hardness significantly. [17] proposed a method to automatically adjust the projection threshold by detecting the surrounding pixel density, [15, 16] also proposed a method to combine with a Chinese character recognitionizer to modify the threshold dynamically to achieve more accurate results. Nevertheless, these approaches are still limited by a specific text format and require many hyperparameters to be tuned manually.

### B. Maintaining the Integrity of the Specifications

Another approach to character segmentation are methods based on deep neural networks. Thanks to the outstanding performance of CNN in image processing in recent years, which greatly reduces the complexity of feature extraction in the past. Many studies have also applied this technique for the segmentation of historical documents[18,19] and have achieved impressive results. However, for the digitization of Chinese historical documents, besides the lack of labeled data, a highly complex language like Chinese with thousands of forms, resulting in related researches, usually can only be applied to specific domains, where requires a series of procedures to classify the results after segmentation further to produce the bounding box[4] for individual character. The reason for not directly segmenting the documents into individual characters is that in the past, to get a better effect on this practice, it usually needs to generate a massive number of ROIs before filtering them one by one through other models in order to form a final bounding box for each character. Nevertheless, it is very inefficient and difficult for dense objects with varying sizes. The details are explained in the next section.

---

[1] A set of rectangle boxes around the region of interest.
[2] A point that can represent the desired object; in this research, we use the centroid of the word as our keypoint.

[3] Intersection over Union is an evaluation metric used to measure the accuracy of an object detector on a particular dataset, which can be formulated as $IoU = \frac{|A \cap B|}{|A \cup B|}$, where $A$ and $B$ are the region of the prediction and ground truth bounding boxes.

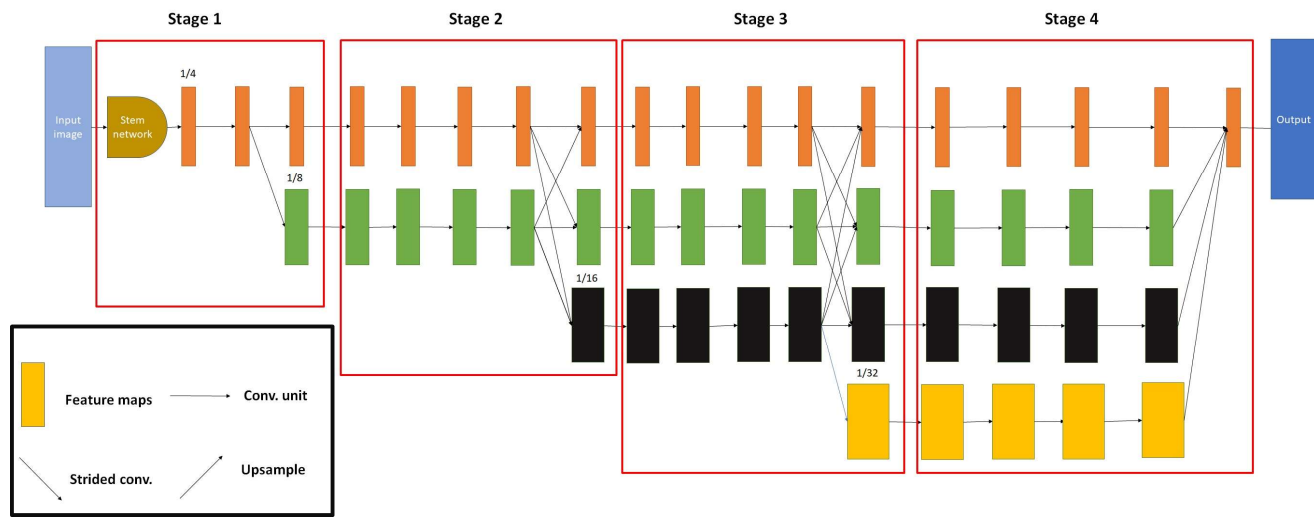

Fig 3. The architecture of our model.

The subject of object detection can be divided into the following two categories, mainly depending on the difference between implementation:

*1) Anchor-based Approach*

The most popular and most effective segmentation model is usually based on anchor-based models, such as the well-known R-CNN families[1-3, 5] or YOLO[4]. These methods use a model that generates a number of anchors of predefined size for the input image and then determines if the proportion of desired objects in each anchor exceeds a predefined threshold. The filtered anchors are then regressed to the desired size and fed into another model to generate the bounding boxes for the desired objects. To achieve high accuracy, these methods usually require the generation of a large number of anchors to obtain a high IoU, and other hyperparameters like the aspect ratio of the anchor box need to be set manually. For historical documents, especially those with high character density or different character sizes, it is often impossible to accurately frame the correct bounding box, and the computational burden is enormous. This is one of the reasons why the anchor-based model did not perform well for this kind of tasks in the past.

*2) Anchorless Approach*

An anchorless approach[6-8] has been proposed in recent years to address the computational burden and generality issues included in the anchor-based approaches. Unlike the anchor-based model, the anchorless approach represents each desired object as a combination of one or more keypoints rather than an anchor. In [6], the model outputs a heatmap representing top-left keypoints and bottom-right keypoints, respectively, and an embedding vector[20] between each set of keypoints will be generated to denote the similarity of the different keypoints to construct the bounding box of the desired object. In [8], in addition to the top-left and bottom-right keypoints, the model also needs to output the estimation of the center keypoint of the desired object, with more accurate predictions. In [7], the model only uses the center of the bounding box as the keypoint to detect the objects and to generate the bounding box, other related attributes such as the length and width of the bounding box need to be derived from the center point. These techniques turn the past object detection problem into a standard keypoint detection problem, greatly reducing the previous anchor-based methods' complexity and making it possible to apply instance segmentation to retrieve characters from the documents directly.

*C. Multi-Scale Feature Representations*

When using CNN in the topic of character segmentation, in addition to a learning method that can be applied to this kind of problem, how to preserve the consistency between high-level and low-level features during the process of feature extraction is one of the most important for position-sensitive tasks like here. In the past studies on image processing, most of them follow the framework of encoder-decoder[27], which gradually decreases the input resolution to extract features to form a high resolution to low-resolution stream. They then use the final low-resolution representation for further processing. To accomplish the related topics of segmentation, it is often necessary, according to past research, to restore the low-resolution representation exported from a classification-like network to the high-resolution representation to get information about the input image, such as the location of different categories in the high-resolution representation, etc. The disadvantage of this strategy is that some spatially precise features are easily lost in the process of conversion to a low-resolution representation, which fails to recover the complete feature in the later process. Thus, how to extract the feature while maintaining the high-resolution representation has been an essential topic for researchers.

[9] as one of the best performing networks on segmentation for two main reasons. The first one is different from the previous encoder-decoder like models, [9] proposes a parallelized structure so that the network can maintain a high-resolution representation while extracting features, instead of trying to recover from the low-resolution representation. The second is that to make the restoration process not distort the original image as much as possible, previous studies usually end up concatenating or adding low-level and high-level representations of the same scale. But in [9], thanks to the parallelization structure, making it able to continuously fuse different features in different scales, such that every feature can be extracted effectively. These reasons make [9] not only semantically strong but also spatially precise.

---

[4] A rectangle box contains the desired object.

## III. APPROACH

In order to construct an efficient and more general character segmentation model with high precision, we propose a network based on [9] for feature extraction and combine with the learning method proposed by [7] to implement an end-to-end trainable anchorless object detection model with high precision, as illustrated in Figure 3.

### A. Problem Formulation

Unlike the anchor-based models used in the past for object detection, which learn further to partition the bounding box from the retrieved ROIs, the anchorless model proposed by [6-8] is mainly based on the model learning the mapping between the input image and the heatmap representing the keypoints. In our model, we do not use the method in [6, 8] that represents the desired objects with multiple keypoints, then uses the embedding vectors to group the different keypoints into the same group to form the final bounding box[20]. Instead, we follow the method proposed by [7], representing each desired object with a single keypoint, the center of the object, then the model predicts the heatmap and related information such as the height and width directly to generate the bounding box. The reason for using the method in [7] instead of [6, 8] is that although [8] used to outperform [7] on average in every aspect, however, we hypothesis that for tasks like in Chinese historical documents, where each Chinese character is very close to each other, and the features of these characters are very similar to one another, is not suitable for the model of [6, 8], which requires to predict the similarity between different keypoints by the embedding vectors. Besides, if we are not predicting a specific character but all Chinese characters in a single page, using the embedding vector method may cause the model to fail to converge the bounding box to the individual Chinese character size. Therefore, in this study, we translate the topic of character segmentation into a standard single keypoint detection problem. The heatmap is an image composed of the probability values representing the centroids in each pixel. In addition to the heatmap, the model should also output the relevant properties for each center point, including the corresponding length and width of each center point to form the binding box, which will be explained later.

### B. Model Details

The backbone of our model is inspired by the architecture of HRNet[9]. The network consists of two components mainly, parallel multi-resolution convolutions and reiterated multi-resolution fusions. With these two components, our model can achieve highly accurate and precise prediction results. Our model's architecture is illustrated in Figure 3.

Firstly, to lessen the computational resources required by the model, the input image first goes through a stem that consists of two stride-2 3x3 convolutions to reduce the resolution to 1/4 of the original one, then enters the stage of parallel multi-resolution convolutions. In this stage, the input will pass through a series of building blocks or bottlenecks from ResNet[11] to extract the features, and finally, a stride-2 3x3 convolution will be added to form a new branch, the resolution of this branch is half of the original input resolution, and the stages of parallel multi-resolution convolutions end here. These branches will be added as the network expands, and will exist in parallel, preserving high-level and low-level representations simultaneously. After that, the parallel branches will enter the multi-resolution fusions stage, where the feature map between different branches will be upsampled, downsampled, or remain unchanged according to the resolution of the main branches, then combine them together to achieve the effect of exchanging information between different resolutions. The entire network is composed of these two components mainly, which can be repeated several times, depending on the size of the model. In our network, we repeat this process three times, including the pre-processing phase; the whole model can be divided into four stages. The first stage decreases the resolution to 1/4 from the original input image alongside with four bottlenecks from ResNet[11], where each block has the number of channels equal to 32, then a new branch is formed via a stride-2 3x3 convolution. Stages 2, 3, and 4, are repeat operations of parallel multi-resolution convolutions combined with multi-resolution fusion, which contain 1, 4, and 3 multi-resolution blocks, respectively. Each multi-resolution block contains 4 building blocks from ResNet[11], and the number of channels in each block is 2C, 4C, 8C, we set C to 32 in our model.

In the final output of the model, simply fuse different branches again and take the highest resolution branch after a 1x1 convolution, and a sigmoid function will be the final prediction result. In order to simplify the model, in addition to the heatmap, which can be represented in probability format, we divide the length and width of the bounding box by the size of the output image to represent them in a probabilistic way.

### C. Objective Functions

In our model, we aim to predict the heatmap $\hat{H}_{xy} \in [0, 1]^{H*W}$ representing the center of each character with height $H$ and width $W$, where $\hat{H}_{xy} = 1$ corresponds to the ground-truth keypoint, while $\hat{H}_{xy} = 0$ is the background. But it is very difficult to output a heatmap prediction that exactly matches the ground-truth locations. Plus, even if the keypoint is slightly off, if it is very close to the corresponding ground-truth location, it will still produce a prediction close to the original binding box. So we follow the approach proposed in [6], for each ground-truth keypoint, we will give it a radius around the keypoint, and the model will get less penalty for false predictions within this radius than outside it. Given the radius, for each ground-truth keypoint $p \in R^2$, a heatmap corresponding to the ground-truth keypoint can be transformed by an unnormalized 2D Gaussian $H_{xy} = \exp(-\frac{(x-\tilde{p}_x)^2+(y-\tilde{p}_y)^2}{2\sigma_p^2})$, where $\sigma_p$ is the value that can be adapted by the radius of the error. An intuitive idea to make the model more accurate in keypoint predictions would be to reduce the value of standard deviation in the Gaussian kernel, but in [21], it is pointed out that this makes optimization more difficult and may lead to even worse results. Therefore, in our model, the value of $\sigma_p$ is set to $\frac{1}{10}$ of the desired object's length and width separately, resulting in a probability distribution that is elliptical rather than circular. If the distributions of two objects overlap each other, we take the maximum value between them instead of the average, such that the different Chinese characters can be classified [22].

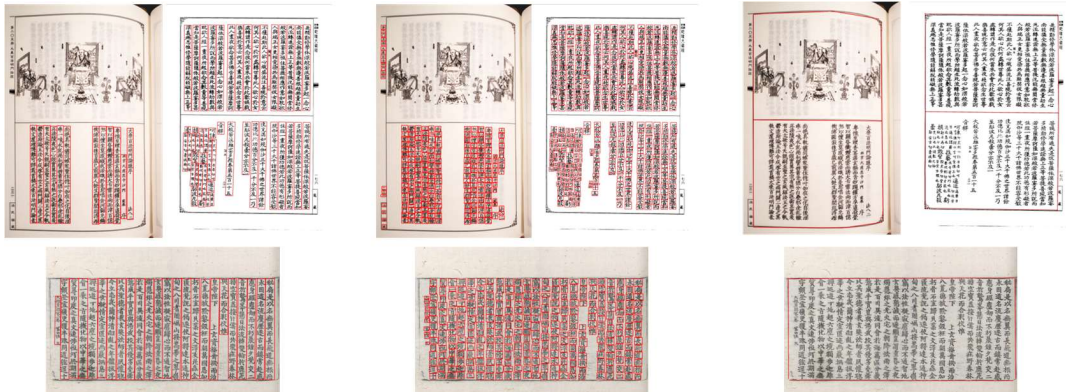

Fig 4. A preview of the dataset MTHv2. The dataset includes three different kinds of annotations, which are line-level annotations, character-level annotations, and boundary line annotations from left to right.

To overcome the problem of imbalanced foreground and background objects, a loss function can be obtained with focal loss[23]:

$$L_h = -\frac{1}{N}\Sigma_{xy}\begin{cases}(1-\hat{H}_{xy})^\alpha \log(\hat{H}_{xy}), H_{xy}=1\\(1-H_{xy})^\beta(H_{xy})^\alpha \log(1-\hat{H}_{xy}), otherwise\end{cases} \quad (1)$$

where $N$ is the number of desired objects in the image, and $\alpha$ and $\beta$ are the hyperparameters for focal loss, here we follow [23] setting $\alpha$ to 2 and $\beta$ to 4 in all our experiments.

In addition to generating a heatmap to represent the center of the Chinese character, to form the bounding box, the model also needs to output each desired object's corresponding length and width. Let $(h_{xy}, w_{xy})$ be the length and width of each desired object in ground-truth keypoint, we use an L1 loss with a mask $M_{xy} \in \{0,1\}^{H*W}$ indicating whether the pixel has a ground-truth keypoint or not to reduce the computational burden, obtaining the following equation:

$$L_s = \frac{1}{N}\Sigma_{xy}|h_{xy} - \widehat{h_{xy}} * M_{xy}| + |w_{xy} - \widehat{w_{xy}} * M_{xy}| \quad (2)$$

To avoid some desired objects are too small, e.g., the desired object's keypoint is between pixels or the inaccurate prediction due to the difference in resolution between input and output. Let $(O_{xy}^1, O_{xy}^2)$ be the offset of the ground-truth keypoint located in $(p_{xc}, p_{yc})$, where $O_{xy}^1 = p_{xc} \bmod 1$, $O_{xy}^2 = p_{yc} \bmod 1$. Similar to Objective (2), an offset loss function can be formulated as:

$$L_{offset} = \frac{1}{N}\Sigma_{xy}|O_{xy}^1 - \widehat{O_{xy}^1} * M_{xy}| + |O_{xy}^2 - \widehat{O_{xy}^2} * M_{xy}| \quad (3)$$

The overall objective functions can be formulated as follows:

$$L_s = \gamma_h L_h + \gamma_s L_s + \gamma_{offset} L_{offset} \quad (4)$$

We set $\gamma_h, \gamma_s, \gamma_{offset}$ to 1, 5, 10 respectively in all our experiments unless specified otherwise.

## IV. EXPERIMENTS

### A. MTHv2 Dataset

One of the biggest problems in deep learning is the preparation of the dataset. The MTHv2 dataset provided by [14] consists of about 3500 images with various types of format. Each image has three different types of annotations: line-level annotations, which contain the position of all text lines in the document and the text content, character-level annotations, which contain the bounding box coordinates of each Chinese character, and the boundary lines annotation, represent the coordinates of different areas of the transcript, such as the coordinates of the illustrations and text content, as shown in Figure 4. In our model, since we will only focus on the segmentation of individual Chinese characters, so we only use character-level annotations.

### B. Training Details

We implement our model in PyTorch[24]. The network is randomly initialized under the default setting of PyTorch. During training, we first set the input image resolution to 512 x 512, which leads to an output resolution of 128 x 128. To reduce overfitting, we adopt random cropping. We use Adam[25] to optimize the Objective (4). We use a batch size of 8 and train the network on 1 Tesla V100 GPU. We train the network for 30 epochs with a learning rate of $1x10^{-6}$. We took a 0.01% of the dataset for testing randomly and the rest for training.

### C. Testing Details

During testing, we first apply non-maximal suppression (NMS) to the output of the model, where we set the confidence score to 0.3 and the IoU threshold to 0.5. Since the resolution of the model output is different from the input image, in order to map the model output back

*Table 1*

Comparisons on MTHv2 dataset. Under the input size 512 x 512, our approach with a small model, trained from scratch, performs better than previous state-of-the-art methods with an acceptable inference time.

| Detection Method | Model | Backbone | Input size | #Params | Inference time(ms) | IoU |
|---|---|---|---|---|---|---|
| **Anchor-based** | EfficientDet[28] | EfficientDet-D5 | 1280x1280 | 34.5M | 70.23 | 0.503 |
| | RetinaNet[11] | ResNet50 | 1024x1024 | 26.3M | 50.34 | 0.257 |
| **Anchorless** | U-Net[10] | ResNet50 | 512x512 | 32.5M | 33.34 | 0.671 |
| | U-Net | ResNet101 | 512x512 | 51.6M | 47.31 | 0.732 |
| | U-Net | MobileNetV2[29] | 512x512 | 8.0M | 31.75 | 0.523 |
| | FPN[13] | ResNet50 | 512x512 | 26.9M | 50.04 | 0.747 |
| | FPN | ResNet101 | 512x512 | 45.9M | 63.21 | 0.761 |
| | FPN | MobileNetV2 | 512x512 | 5.2M | 48.04 | 0.495 |
| | PSPNet[26] | ResNet50 | 528x528 | 3.8M | 22.77 | 0.472 |
| | PSPNet | ResNet101 | 528x528 | 3.8M | 21.49 | 0.234 |
| | PSPNet | MobileNetV2 | 528x528 | 1.6M | 19.53 | 0.323 |
| | ***Ours*** | *HRNetV2-W32[9]* | *512x512* | *9.5M* | *20.41* | ***0.814*** |

to the input, we multiply the height, width, and offset maps output from the model by $\frac{In_h}{Out_h}$ and $\frac{In_w}{Out_w}$ respectively. Where $In_h$ and $In_w$ are the height and width of the input image, $Out_h$ and $Out_w$ are the height and width of the output image. The final results are shown in Figure 5.

*D. Ablation Study*

To proof that our method and backbone are superior compared to the others, we examine our model alongside with previous state-of-the-art models as shown in **Table 1**. All experiments are trained with batch size 8 and 80 epochs. For anchor-based models, we follow the default setting shown in the original paper and show that, compared to the anchor-based detection method, the anchorless approach is more general to the task in character segmentation for historical documents. We also examine other models with different backbones to show that with the parallel architecture, the feature to be preserved can make a significant improvement in the detection results.

V. CONCLUSION

In this research, we propose a character detection model named HRCenterNet combining with an anchorless approach in object detection and a parallelized mechanism so that the global feature and local feature can be considered together, and high-level and low-level representations can interact with each other under multi-scale fusion. The experiments and the results have shown that our model is able to achieve the best speed-accuracy compared to the others.

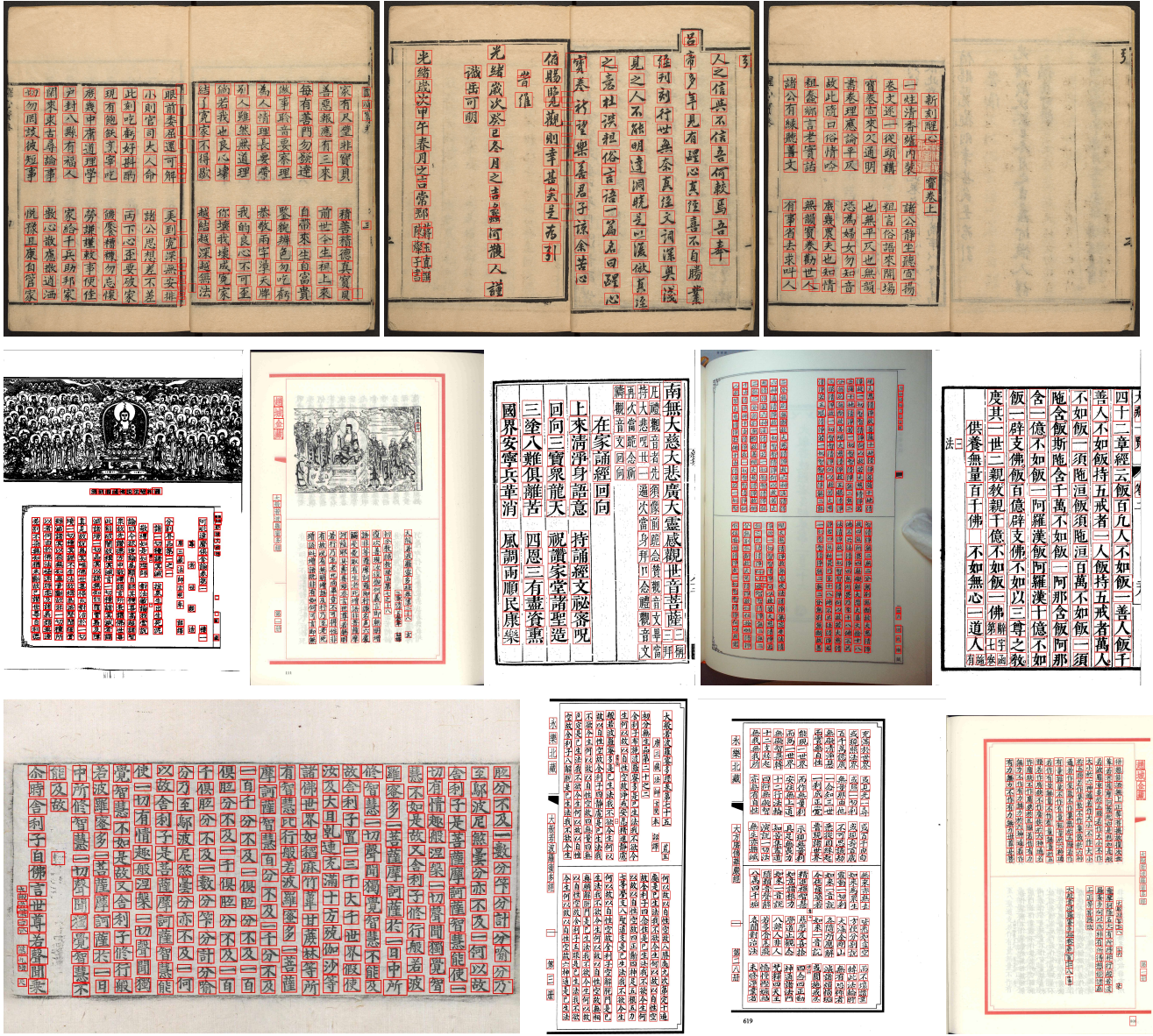

Figure 5. Detection results of our model on the testing data. In addition to processing documents in variable formats, our model can even achieve highly accurate segmentation results in images that contain illustrations or character in the background.

ACKNOLEDGEMENTS

This research was support in part by the contracts MOST-109-2813-C-004-011-E and MOST-107-2200-E-004-009-MY3 of the Ministry of Science and Technology of Taiwan.